\documentclass[11pt,letterpaper]{article}
\usepackage{emnlp2016}
\usepackage{times}
\usepackage{latexsym}
\usepackage{graphicx}
\usepackage{amsmath}
\usepackage{subcaption}
\usepackage{float}
\usepackage{algorithm}
\usepackage{multirow}
\usepackage{array}
\usepackage[noend]{algpseudocode}
\DeclareMathOperator*{\argmin}{arg\,min}

\usepackage{color}

\emnlpfinalcopy

\title{Fill it up: Exploiting partial dependency annotations \\ in a minimum spanning tree parser.}

\author{
 Liang Sun$^1$ \And Jason Mielens$^2$ \\\\
 \begin{tabular}{*{2}{>{\centering}p{.45\textwidth}}}
 $^1$Department of Mechanical Engineering& $^2$Department of Linguistics \tabularnewline
 The University of Texas at Austin & The University of Texas at Austin \tabularnewline
 {\tt sally722@utexas.edu} & {\tt \{jmielens,jbaldrid\}@utexas.edu} \\
\end{tabular} \And Jason Baldridge$^2$}

\date{}

\begin{document}
\maketitle
\begin{abstract}
Unsupervised models of dependency parsing typically require large amounts of clean, unlabeled data plus gold-standard part-of-speech tags.  Adding indirect supervision (e.g. language universals and rules) can help, but we show that obtaining small amounts of direct supervision---here, partial dependency annotations---provides a strong balance between zero and full supervision. We adapt the unsupervised ConvexMST dependency parser to learn from partial dependencies expressed in the Graph Fragment Language. With less than 24 hours of total annotation, we obtain 7\% and 17\% absolute improvement in unlabeled dependency scores for English and Spanish, respectively, compared to the same parser using only universal grammar constraints.
\end{abstract}

\section{Introduction}

Unsupervised parsing solutions are simultaneously an attractive yet troublesome method for handling low-data scenarios. The performance of unsupervised parsers has increased dramatically in recent years \cite{klein:2004,naseem:2010}, making them a potentially viable option for constructing labeled corpora on limited budgets. However, their performance is often outmatched by small amounts of labeled data \cite{blunsom:2010,spitkovsky:2012}. Further, recent work using linguistically-informed error analysis on unsupervised Combinatory Categorial Grammar parsing shows that entire syntactic phenomena are outside the scope of existing unsupervised parsers \cite{bisk:2015}. Accordingly, most recent work in this area has focused on methods of providing sources of indirect annotation, whether via linguistic world-knowledge \cite{naseem:2010,grave-elhadad:2015}, partial annotations \cite{flannery:2011,mielens-sun-baldridge:2015} or cross-lingual information transfer \cite{naseem:2012}.

With unsupervised parsing, data collection is not entirely eliminated: a large amount of clean, relevant data is needed. Also, evaluations of unsupervised techniques typically rely on gold part-of-speech tags. Obtaining clean data for many languages is actually a difficult process--complicated by issues such as language identification, digitization, and varying or absent orthographies. This challenge also exists in many domain adaptation scenarios.

We explore the effectiveness of creating small amounts of labeled data using the Graph Fragment Language (GFL), an annotation scheme designed for speed and ease \cite{schneider:2013,gflweb:2014}. We create 270 English and 2297 Spanish partial sentence annotations using GFL, using a mix of expert and non-expert annotators. We then adapt the minimum spanning tree based parsing technique of Grave \& Elhadad \shortcite{grave-elhadad:2015} to use these partial annotations in addition to universal dependency rules it already exploits. Throughout this work we will refer to this parser as ConvexMST.\footnote{Code available at github.com/jmielens/convex-mst}

We present parsing results with and without gold part-of-speech tags. When using predicted POS tags, our experiments show that exploiting cheap, incomplete direct supervision in addition to language universals provides large absolute performance improvements for both English and Spanish: 6.3\% for the former and 17.3\% for the latter. Furthermore, the ConvexMST parser dramatically outperforms the Gibbs sampler parser of Mielens et al. \shortcite{mielens-sun-baldridge:2015} using the supervision (English: +5.2\%; Spanish: +14.4\%). We also show that the extra supervision provided by gold POS tags heavily influences results; in particular, it inflates performance when only using language universals. Experiments that rely on gold POS tags alone are thus not reliable indicators of performance in true low-resource settings.

\section{Annotation}

\subsection{Graph Fragment Language}

\begin{figure}[t]
  \centering
    \includegraphics[width=0.5\textwidth,trim={0 0 0 0}]{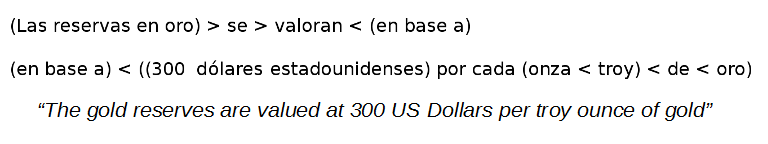}
    \caption{Spanish GFL Example: Parentheses indicate a constituent-style bracket, angle brackets indicate direct dependency relations.}
    \label{fig:gfl-example}
\end{figure}

We use the Graph Fragment Language (GFL) \cite{schneider:2013} to allow for light-weight, simple annotations that our annotators can easily learn and use confidently. The choice of annotation scheme is particularly important: we seek to optimize annotation speed rather than full-specification or high accuracy. In previous studies, the use of GFL has allowed for annotation rates of 2-3 times that of traditional dependency annotations while still maintaining a useful level of annotation density \cite{gflweb:2014,mielens-sun-baldridge:2015}.
Hwa \shortcite{hwa:1999} demonstrated that it is most effective to provide high-level sentence constituents to a parser and allow it to fill in the low-level information itself.

The GFL annotation in Figure~\ref{fig:gfl-example} shows two distinct notations. Constituent brackets are specified by parentheses and direct dependencies by angle brackets. Many words and phrases are underspecified. Allowing partial annotations dramatically increases the speed at which annotators can work, while simultaneously reducing error rates. These two effects both arise from being able to leave difficult or tedious portions of a sentence unspecified.

\subsection{Filling in Partial Dependencies}

A partial annotation produces a set of dependency tree fragments. Compared to an unlabeled sentence, this can substantially reduce the work a parser must do. When working with partial dependencies, there are two paths that can be taken with regard to overall model-building. In a `Fill-then-Parse' setup, the partial dependencies are first filled-in to produce full dependencies that are then used to train a standard dependency parser. In a `Fill+Parse' setup, one model both fills in and parses new sentences. 

We use a Fill+Parse setup, while previous work focused on Fill-then-Parse. The major benefit of the former is that learning can be sensitive to the source of an arc in the training data---e.g., whether it came from an annotator or a universal rule. Fill-then-Parse obscures this distinction and not knowing how trustworthy an arc is can lead to additional errors. Indeed, Fill+Parse method produces better results for our datasets than Fill-then-Parse (see Section~\ref{ssec:parsing-results}).

\subsection{Simulated Cost Comparison}
\label{ssec:cost}

Many factors influence the cost of creating a corpus. Our goal is to minimize cost relative to the performance of a parser trained with the corpus. The actual cost of finding and paying annotators is the most obvious factor, and it will typically be higher for a low-resource language or highly specialized domain.
Using a light-weight partial annotation scheme like GFL has the potential to increase the pool of qualified annotators and alleviate this challenge.

\begin{figure*}[t]
  \centering
  \begin{subfigure}[b]{0.48\textwidth}
    \includegraphics[width=\textwidth]{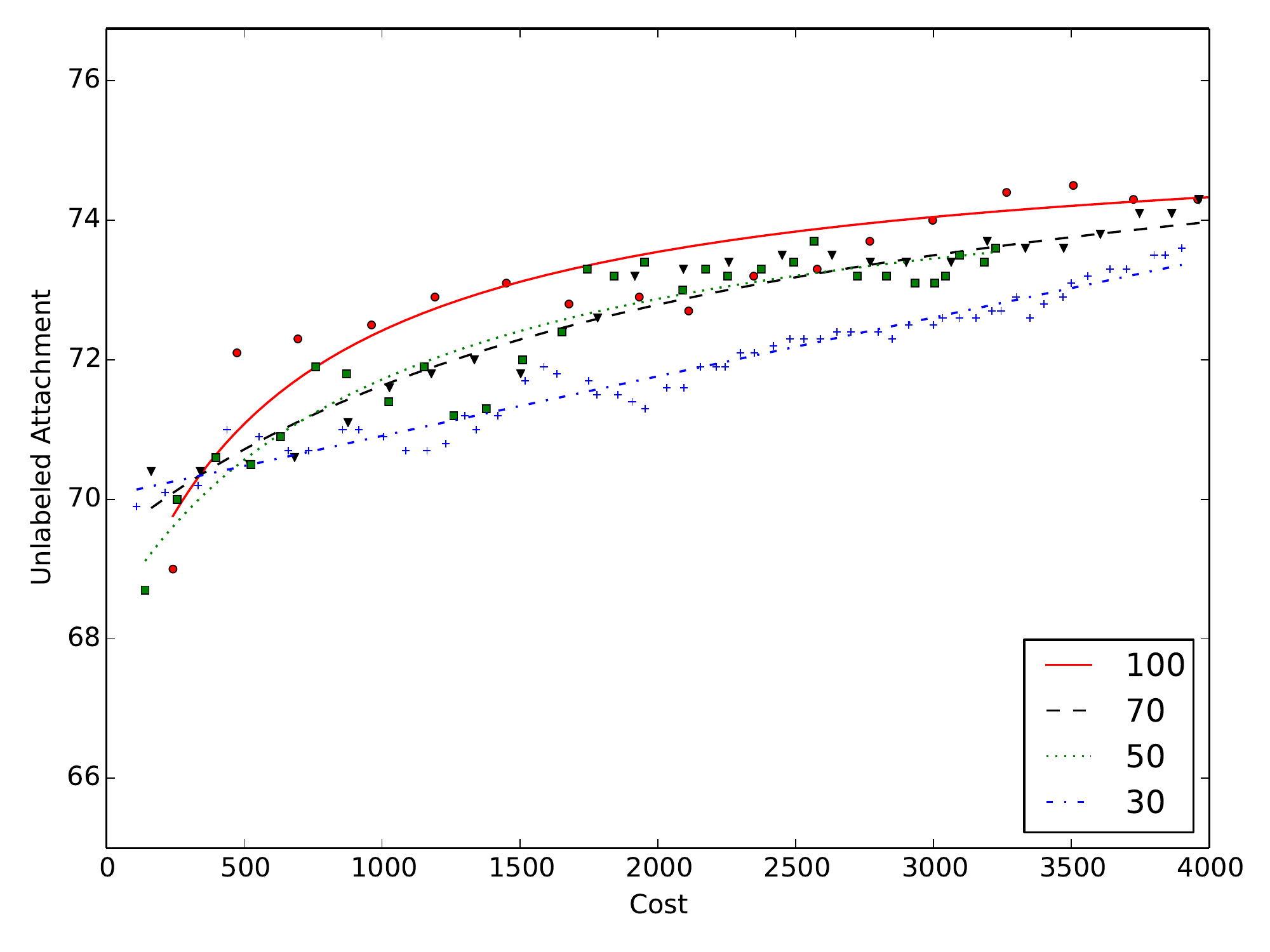}
    \caption{Equal Cost}
    \label{fig:equal-cost}
  \end{subfigure}
  \quad
  \begin{subfigure}[b]{0.48\textwidth}
    \includegraphics[width=\textwidth]{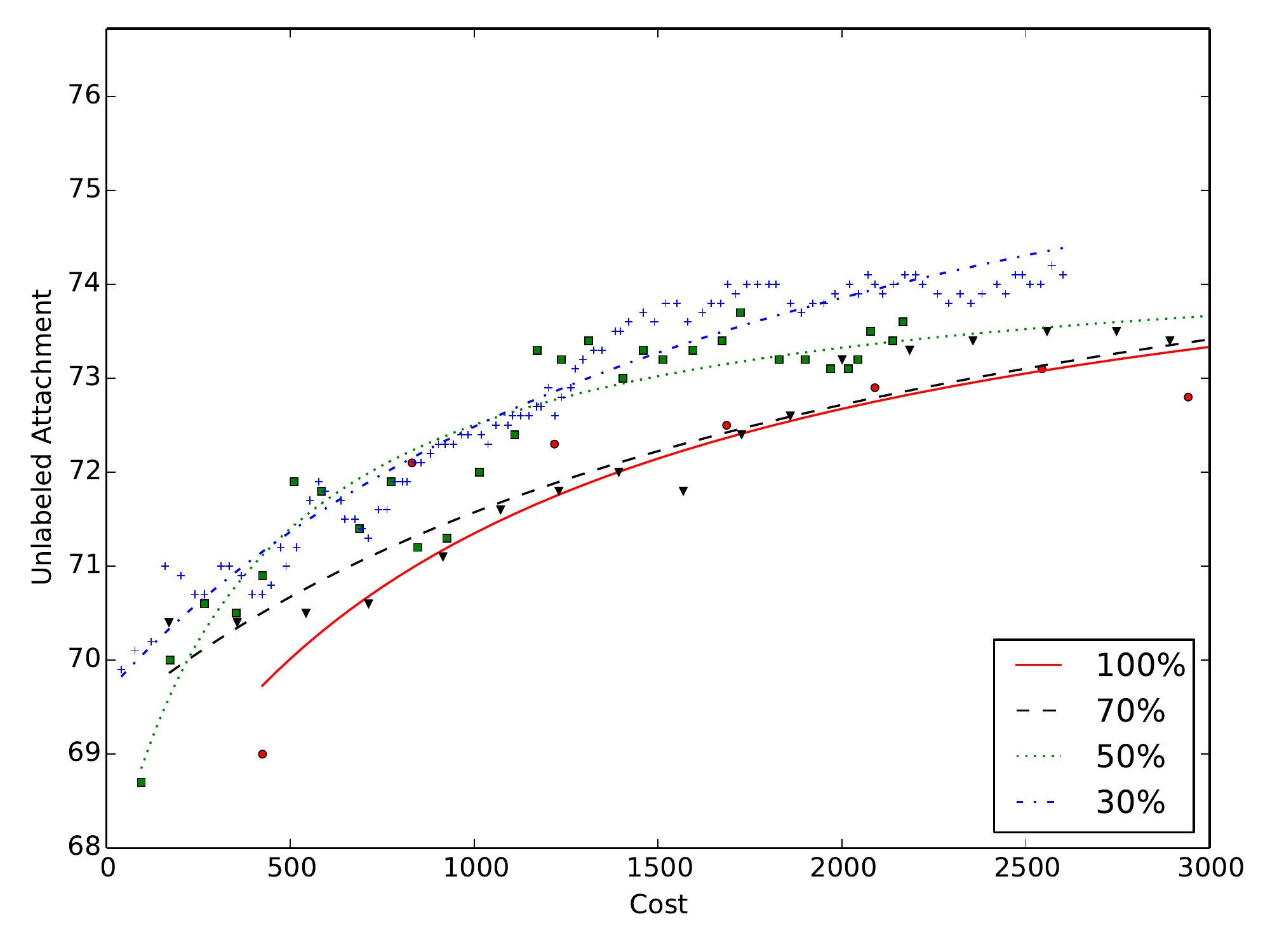}
    \caption{Variable Cost}
    \label{fig:variable-cost}
  \end{subfigure}
  \caption{Comparison of performance versus total cost, Equal Cost is the sum of all specified dependencies, Variable Cost weights dependencies by completion percentage. Run on Spanish data using simulated partial dependencies.}
  \label{fig:cost-comparison}
\end{figure*}

Given a partial annotation scheme like GFL, an additional cost factor is that of obtaining a particular level of completion for each sentence. Consider that for any sentence there are both `low-hanging fruit' dependencies such as determiner attachment, and more difficult dependencies such as preposition attachment and long-distance relations. Harder dependencies take longer to annotate (and thus cost more), so it is worth considering cost metrics that incorporate completion percentage. In the absence of timing/expense data, we can simulate this intuition with a variable cost model for which each an additional dependency annotated in a sentence is more expensive than the previous one.

Figure \ref{fig:cost-comparison} demonstrates the impact of completion cost. Parsing accuracies (for our parser introduced in the next section) are shown at different costs, under (a) simple equal (per arc) cost and (b) variable cost. We simulated the construction of various corpora by deriving partial dependencies from gold standard annotations), and show the cost curves for different sentence completion rates. 100\% completion produces the best performance with equal costs, but under the more realistic variable cost model, 30\% and 50\% completion win. We show later that this pattern holds under actual timed annotation.

Garrette \shortcite{garrette:2015} demonstrated the benefit of partial annotations for CCG parsing. They focused on the number of (partial) bracket annotations (as a proxy for annotation time), holding this fixed while varying the number of sentences. Strikingly, they found that having 40\% of brackets across the full dataset was better than full brackets for 80\% of the corpus. This result uses an equal cost-per-bracket assumption, so the difference would be even more favorable to partial annotations with a variable cost.

\subsection{Unsupervised vs. Partial Annotations}

Without any direct annotations, we must rely on indirect supervision such as universal grammar rules, cross-lingual information transfer, and domain adaptation. Following Grave \& Elhadad \shortcite{grave-elhadad:2015}, we use the universal grammar rules in Table \ref{tab:ug-rules}. Indirect supervision via these rules is achieved by biasing produced trees to conform to the rules. This is the only form of dependency supervision considered by Grave \& Elhadad, though they do provide additional direct supervision via gold part-of-speech tags.

\subsection{Data}
\label{ssec:data}

We use two sources of data. To compare with prior work, we use the universal treebanks (version 2.0), which cover ten languages from a variety of language families \cite{mcdonald:2013}. We obtained GFL annotations for a subset of the English data, originally from WSJ Section 03 of the Penn Treebank, and we use simulation techniques to produce partial dependencies for the other languages.

\begin{table}
    \centering
    \small
    \begin{tabular}{cc}
    \hline
    Verb $\mapsto$ Verb & Noun $\mapsto$ Noun \\
    Verb $\mapsto$ Noun & Noun $\mapsto$ Adj  \\
    Verb $\mapsto$ Pron & Noun $\mapsto$ Det  \\
    Verb $\mapsto$ Adv  & Noun $\mapsto$ Num  \\
    Verb $\mapsto$ Adp  & Noun $\mapsto$ Conj \\ \hline
    Adj $\mapsto$ Adv   & Adp $\mapsto$ Noun  \\
    \end{tabular}
    \caption {Universal Grammar Rules}
    \label{tab:ug-rules}
\end{table}

Our second data source is the Spanish dependency treebank from the AnCora corpus \cite{taule:2008}.
For 1410 unique sentences of AnCora, we have partial dependencies specified in GFL by twelve annotators. Most sentences received a single partial annotation from a single annotator, but one section of the corpus was annotated by all annotators. As the original corpus is fully-specified for gold dependencies, we can measure annotator agreement with a gold standard.

The background and experience of the annotators varied considerably. Roughly one third were native Spanish speakers, with the rest ranging from fluent non-native speakers to a few with just a single year of formal study. This was done intentionally to provide a large variance in the types and quality of annotations that they were able to provide.

Each annotator was trained for just 30 minutes. The nature of the annotations was explained and a small number of guidelines were provided. For instance, annotators were told that typically adjectives are dependents of nouns, nouns are dependents of verbs, and so on. These guidelines amounted to a summary of the rules in Table~\ref{tab:ug-rules}. During the annotation sessions, annotators were told to ask as many clarifying questions as needed, although in practice they needed very little guidance. Post-experiment debriefing interviews suggested that the straight-forward nature of the GFL notation was very helpful and became clear within a few example sentences.

Despite minimal training time, annotators were able to produce relatively consistent annotations that agreed in large part with other annotators. Table \ref{tab:agreement} shows both pair-wise and overall agreement between annotators when considering arcs that each of the annotators in the pair had provided a head for. Overall agreement was high, with most pairwise numbers in the 70-80's, and agreement for individual annotators to the group is even higher -- mostly in the 80's. 

The partial annotation task proved helpful in terms of speed; our annotators were able to cover 750 tokens/hr, which compares favorably to the processes of the Penn Treebank, which achieved rates of 750-1000 tokens/hr for English \cite{marcus1993}, and 300-400 tokens/hr for Chinese \cite{xue2005}, both making use of initial parse suggestions from an existing parser. Efforts not using an existing parser proceed even slower; for instance the Ancient Greek Dependency Treebank reported rates of 100-200 tokens/hr \cite{bamman2011}.

\begin{table}
    \resizebox{.5\textwidth}{!}{\begin{tabular}{l|c|c|c}
    ~                      & Partial EN & Full ES & Partial ES \\ \hline
    Unique Sentences       & 270          & 135       & 1410          \\
    Total Sentences        & 270          & 135       & 2162          \\ \hline
    Number of Annotators   & 2          & 1       & 12         \\
    Total Annotation Hours & 8          & 13      & 72         \\
    \end{tabular}}
    \caption {Training Set Statistics}
    \label{tab:training-stats}
\end{table}

\begin{table*}[t]
\centering
\footnotesize
 \resizebox{.75\textwidth}{!}{\begin{tabular}[b]{c|c|c|c|c|c|c|c|c|c|c|c|c|c}
    Annotator     & 1    & 2    & 3    & 4    & 5    & 6    & 7    & 8    & 9    & 10   & 11   & 12   & Avg. \\ \hline
    1     &  1   & 0.73 & 0.9  & 0.88 & 0.55 & 0.77 & 1    & 0.94 & 0.9  & 0.28 & 0.95 & 0.67 & .80  \\
    2     & 0.73 & 1    & 0.78 & 0.83 & 0.95 & 0.62 & 0.77 & 0.75 & 1    & 0.27 & 0.8  & 0.85 & .78  \\
    3     & 0.9  & 0.78 & 1    & 0.85 & 0.64 & 0.8  & 0.9  & 0.82 & 0.96 & 0.3  & 0.85 & 0.72 & .79  \\
    4     & 0.88 & 0.83 & 0.85 & 1    & 0.6  & 0.88 & 0.83 & 0.92 & 1    & 0.33 & 0.91 & 0.68 & .81  \\
    5     & 0.55 & 0.95 & 0.64 & 0.6  & 1    & 0.46 & 0.64 & 0.55 & 0.83 & 0.23 & 0.59 & 0.85 & .66  \\
    6     & 0.77 & 0.62 & 0.8  & 0.88 & 0.46 & 1    & 0.88 & 0.74 & 0.88 & 0.16 & 0.75 & 0.6  & .71  \\
    7     & 1    & 0.77 & 0.9  & 0.83 & 0.64 & 0.88 & 1    & 0.94 & 1    & 0.2  & 1    & 0.67 & .82  \\
    8     & 0.94 & 0.75 & 0.82 & 0.92 & 0.55 & 0.74 & 0.94 & 1    & 1    & 0.36 & 0.94 & 0.7  & .81  \\
    9     & 0.9  & 1    & 0.96 & 1    & 0.83 & 0.88 & 1    & 1    & 1    & 0    & 1    & 0.81 & .87  \\
    10    & 0.28 & 0.27 & 0.3  & 0.33 & 0.23 & 0.16 & 0.2  & 0.36 & 0    & 1    & 0.11 & 0.12 & .28  \\
    11    & 0.95 & 0.8  & 0.85 & 0.91 & 0.59 & 0.75 & 1    & 0.94 & 1    & 0.11 & 1    & 0.68 & .80  \\
    12    & 0.67 & 0.85 & 0.72 & 0.68 & 0.85 & 0.6  & 0.67 & 0.7  & 0.81 & 0.12 & 0.68 & 1    & .70  \\ \hline
    Total & .85  & .86  & .86  & .9   & .82  & .77  & .98  & .92  & .96  & .61  & .94  & .82  & ~    \\
    \end{tabular}}
    \caption {Pair-wise and total agreement by annotator. The `Total' row shows agreement with the set of all other annotators and the `Avg.' column is the average pairwise agreement.}
   \label{tab:agreement}
\end{table*}

\subsection{POS-Tagging}
\label{sssec:pos}

Our goal is to minimize real-world costs associated with producing a finished parsing model. To this end, we trained our own POS taggers using type label annotations \cite{garrette:2013} rather than using gold-standard tags. We use universal POS tags rather than the finer-grained sets the source corpora use, both for simplicity and cross-language comparisons \cite{petrov:2011}.

We trained taggers for all languages using a limited amount of the available gold data---ensuring that the accuracy is comparable with low-resource human-sourced taggers. We extract types from the corpus, rank them by frequency, and take the most frequent types to train the tagger. The cutoff on how many types to take is derived from the number of types the annotators in Garrette et al. \shortcite{garrette:2013} were able to produce in two hours. The taggers all obtain around 80\% accuracy.

\section{Method}

\subsection{Convex-MST}

This section provides a brief overview of the core parsing algorithm; for full details, see Grave \& Elhadad \shortcite{grave-elhadad:2015}. We begin by considering a binary vector $\mathbf{y}$ that encodes all of the dependencies in our corpus, such that $\mathbf{y}_{ijk} = 1$ if sentence $i$ has an arc with dependent $j$ and head $k$. This representation leads to the problem formulation in Equation \ref{eq:ge-prob}, where $Y$ is the convex hull of all the valid tree assignments for $\mathbf{y}$, $\mathbf{n}$ is the number of possible dependency arcs in the corpus, $\mathbf{u}$ is a penalty vector that penalizes potential dependency arcs that are not in the set of universal dependency rules, and $\mathbf{w}$ is a weight vector learned during training

\begin{equation}
\label{eq:ge-prob}
\min_{\mathbf{y} \in Y} \min_{\mathbf{w}} \frac{1}{2n} \lVert \mathbf{y} - \mathbf{Xw} \rVert^{2}_{2} + \frac{\lambda}{2} \lVert \mathbf{w} \rVert^{2}_{2} - \mu \mathbf{u}^{T} \mathbf{y}
\end{equation}

\noindent
This problem can be solved using Algorithm \ref{alg:optimization} \cite{grave-elhadad:2015}.

\begin{algorithm}[t]
\caption{Optimization algorithm from Grave \& Elhadad (2015)}\label{alg:optimization}
\begin{algorithmic}[1]
\For{$r\not=0$}

Compute the optimal $\mathbf{w}$:

$\mathbf{w}_{t} = \argmin_{\mathbf{w}} \frac{1}{2n} \lVert \mathbf{y}_{t} - \mathbf{Xw} \rVert^{2}_{2} + \frac{\lambda}{2} \lVert \mathbf{w} \rVert^{2}_{2}$

Compute the gradient w.r.t. $\mathbf{y}$:

$\mathbf{g}_{t} = \frac{1}{n} (\mathbf{y}_{t} - \mathbf{Xw}_{t}) - \mu \mathbf{u}$

Solve the linear program:

$\mathbf{s}_{t} = \min_{\mathbf{s} \in Y} \mathbf{s}^{T}\mathbf{g}_{t}$

Take the Franke-Wolfe step:

$\mathbf{y}_{t} = \gamma_{t}\mathbf{s}_{t} + (1 - \gamma_{t})\mathbf{y}_{t}$

\EndFor
\end{algorithmic}
\end{algorithm}

\subsection{Partial Dependency Features}
\label{ssec:partial-features}

The main  modification we make is to add an additional term to penalize arcs that disagree with partial annotations. Let $\mathcal{S}$ be the set of all indices on $\mathbf{y}$ where that head-dependent pair conforms to one of the universal rules. Then we can require that some proportion of the arcs in the corpus satisfy:

\begin{equation*}
\frac{1}{n}\sum_{i \in \mathcal{S}}y_i \ge c
\end{equation*}

\noindent
This is equivalent to $\mathbf{u}^{T}\mathbf{y} \ge c$, where:

\begin{equation*}
u_{i}=\begin{cases}
    1/n, & \text{if $i \in \mathcal{S}$}.\\
    0, & \text{otherwise}.
  \end{cases}
\end{equation*}

\noindent
This is how the penalty term $\mu \mathbf{u}^{T} \mathbf{y}$ from (\ref{eq:ge-prob}) is derived. Similarly, we can add another penalty term that ensures a certain percentage of the arcs conform to the arcs specified by annotators. If we let $\mathcal{G}$ be the set of all indicies on $\mathbf{y}$ where the word pair conforms to the GFL annotations, then it is simple to construct an additional penalty term $\xi \mathbf{v}^{T} \mathbf{y}$.

There is a slight difference between the GFL penalty term and the universal rule penalty term. Whereas the universal rule penalty is based simply on whether the arc conforms or does not conform to the rules, the GFL annotations naturally lead to a three-way distinction: the annotation can specify that an arc \textit{should} be present, \textit{should not} be present, or make no commitment. 

Accordingly, we modify $\mathcal{G}$ to be two sets, $\mathcal{G}_w$ and $\mathcal{G}_b$, where $\mathcal{G}_w$ is the set of all indicies on $\mathbf{y}$ where the word pair should have an arc, and $\mathcal{G}_b$ is the set of all indicies on $\mathbf{y}$ where the word pair should \textit{not} have an arc. We refer to these as the whitelist and blacklist, accordingly. Under this formulation, the GFL-based penalty term $\xi \mathbf{v}^{T} \mathbf{y}$ is now made with:

\begin{equation*}
v_{i}=\begin{cases}
    1/n, & \text{if $i \in \mathcal{G}_w$}\\
    -1/n, & \text{if $i \in \mathcal{G}_b$}\\
    0, & \text{otherwise}
  \end{cases}
\end{equation*}

\noindent
This leads to the modified objective function in (\ref{eq:gfl-obj}), which now seeks to find a solution that minimizes the number of arcs that violate both universal rules and the annotator-specified fragments.

\begin{equation}
\label{eq:gfl-obj}
\min_{\mathbf{y} \in Y} \min_{\mathbf{w}} \frac{1}{2n} \lVert \mathbf{y} - \mathbf{Xw} \rVert^{2}_{2} + \frac{\lambda}{2} \lVert \mathbf{w} \rVert^{2}_{2} - \mu \mathbf{u}^{T} \mathbf{y} - \xi \mathbf{v}^{T} \mathbf{y}
\end{equation}

\noindent
When no GFL annotations are specified for the corpus, the GFL penalty term goes to zero and the objective function reverts to its original formulation.

\begin{figure}[t]
  \centering
    \includegraphics[width=.5\textwidth]{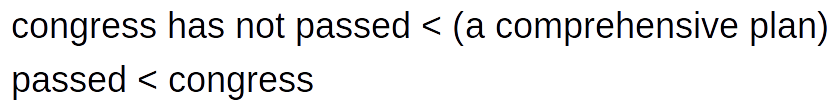}
    \caption{GFL Whitelisting vs. Blacklisting}
    \label{fig:white-black}
\end{figure}

Specific arcs are added to $\mathcal{G}_w$ and $\mathcal{G}_b$ in a number of ways, based on the different types of GFL annotation. Consider the GFL annotation in Figure~\ref{fig:white-black}. Here, the annotator has specified a direct dependency with `passed' as the head of `congress'. The arc `passed $\leftarrow$ congress' is added to $\mathcal{G}_w$, while all other arcs of the form `$X \leftarrow$ congress' are added to $\mathcal{G}_b$ because `congress' may only have a single head.

Brackets may also result in additions to the whitelist and blacklist. In Figure~\ref{fig:white-black}, `a comprehensive plan' is bracketed. In this case, no arcs can be whitelisted, but many can be blacklisted. For instance, no word external to the bracket may be headed by a word in the bracket. This means arcs such as `plan $\leftarrow$ congress' must be in $\mathcal{G}_b$. 

Also, `passed' is indicated as the head of the entire bracket. We cannot whitelist any specific arcs with this information (since we do not know the head of the bracketed expression), but we know that no word internal to the bracket is headed by any word external to it, other than `passed'. Hence, arcs such as `congress $\leftarrow$ plan' must be in $\mathcal{G}_b$.

\section{Experiments and Discussion}
\label{sec:results}

We consider both simulated and actual partial annotations. Results based on actual annotation are the most important as they provide our best measure of performance under a realistic annotation setting. However, our Spanish annotators had only six hours each, and there was no inter-annotator communication or creation of annotation conventions, and no attempt to have them adopt the conventions in the gold-standard AnCora dependencies we evaluate against.  Because of this, we include simulation results to eliminate this source of divergence to better measure the effectiveness of different methods for filling in missing arcs in a partial annotation. It of course also allows us to measure this for all the languages in the Universal Dependencies treebanks.

We consider three different supervision settings for ConvexMST:

\begin{itemize}
\item \textit{UG} uses just the universal grammar based features, which is equivalent to the method used by Grave \& Elhadad \shortcite{grave-elhadad:2015}.
\item \textit{GFL} uses just the human specified features.
  
\item \textit{GFL+UG} uses both. 
  
\end{itemize}

\noindent These three methods correspond with $\xi \neq 0$, $\mu \neq 0$, and $\xi  \mu \neq 0$ in Equation 2.  The training sets correspond with the `Partial EN' and `Partial ES' sets from Table \ref{tab:training-stats}. The set of sentences annotated with GFL is used as the training set for the \textit{GFL},  \textit{UG}, and \textit{GFL+UG} methods.

\subsection{Simulated partial dependencies}

\begin{table}[t]
  \centering
  \small
  \resizebox{.35\textwidth}{!}{\begin{tabular}{l|c|c|c}

    & \multicolumn{3}{c}{Degradation} \\ \hline
    Language & None & Light & Heavy \\ \hline
    DE		& 69.6       & 69.5              & 68.2              \\
    EN		& 79.5      & 77.5              & 72.8              \\
    ES		& 78.0      & 76.5              & 71.5              \\
    FR		& 82.6      & 82.3              & 75.7              \\
    IT		& 82.0      & 81.8              & 77.4              \\
    PT-BR	& 80.6	    & 80.0              & 72.4              \\
    SV		& 77.7	& 77.1              & 76.9              \\ \hline
    Average  & 78.5	& 77.8              & 73.6              \\ \hline
  \end{tabular}}
  \caption {Simulated degradation results. \textit{Light} has 40\% of arcs removed and \textit{Heavy} has 70\% removed.}
  \label{tab:degradation}
\end{table}

\begin{figure}[t]
  \centering
  \includegraphics[width=0.48\textwidth]{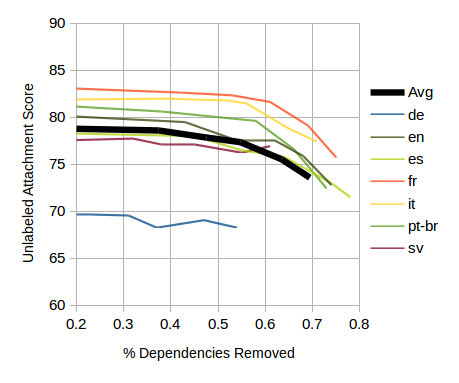}
  \caption{Degradation Simulations}
  \label{fig:degradation}
\end{figure}

Simulated partial dependencies are produced by removing dependencies via a stochastic process that approximates how we instructed human annotators to focus their efforts. Arcs are removed top-down, with arcs lower in the tree being more likely to be deleted. This results in trees with more high-level structures and less lower-level information. Figure~\ref{fig:degradation} demonstrates the stability of our parser under varying levels of such gold tree degradation. Missing arcs were recovered using our parse imputation scheme (using GFL+UG features), and the resulting parser was applied to the evaluation sentences. Accuracy decreases slightly to around 60\% removal, and then degrades more rapidly after that. Table~\ref{tab:degradation} provides numeric data for the simulations.

\begin{table}[t]
    \small
    \centering
    \resizebox{.48\textwidth}{!}{\begin{tabular}{l|l|cc|cc}
    \multirow{2}{*}{Parser} & \multirow{2}{*}{Features}     & \multicolumn{2}{c|}{Gold Tags}  & \multicolumn{2}{c}{Predicted Tags}   \\
    ~ & ~     & EN & ES & EN & ES \\ \hline
    RB       & N/A & 17.1 & 28.0 & 17.1 & 28.0 \\ \hline
    Gibbs    & GFL & 60.2 & 65.3 & 55.8 & 52.7\\ \hline
    \multirow{3}{*}{ConvexMST}       & UG & 63.1 & 63.5 & 56.9 & 50.0 \\
          & GFL & 65.9 & 70.5 & 61.2 & 67.1   \\
       & UG+GFL & \textbf{68.2} & \textbf{71.3} & \textbf{63.2} & \textbf{67.3}  \\
    \end{tabular}}
    \caption {Directed dependency accuracy on English and Spanish universal treebanks using annotator provided GFL annotations, 10 or fewer words.}
    \label{tab:parsing-results}
\end{table}

\subsection{Annotator-sourced partial dependencies}
\label{ssec:parsing-results}

Table \ref{tab:parsing-results} gives semi-supervised parsing results on the English and Spanish treebanks for sentences with 10 or fewer words. To investigate the impact of POS taggers on parsing results, we conducted two series of experiments using POS tags trained by our own tagger as discussed in Section~\ref{sssec:pos} (\textit{Predicted Tag}) and gold POS tags extracted from treebank (\textit{Gold Tag}). We compare against a right-branching baseline and the Gibbs parser of Mielens et al. \shortcite{mielens-sun-baldridge:2015}.

All the parsing methods handily beat the right-branching baseline. ConvexMST-UG (the model of Grave and Elhadad \shortcite{grave-elhadad:2015}) beats the Gibbs parser with gold POS tags, but the ranking switches with predicted POS tags. This shows the effectiveness of ConvexMST, but highlights its brittleness with respect to tagging errors: bad tags lead to poor guidance from language universals. ConvexMST-GFL easily beats both these approaches: it exploits partial annotations much more effectively than the Gibbs parser and learns effectively without language universals. The difference is especially marked for predicted POS tags: ConvexMST-GFL beats ConvexMST-UG by 4.3\% for English and 17.1\% for Spanish. (Recall that there were 8 hours of annotation for English and 72 hours for Spanish.)

\begin{figure}[H]
  \centering
    \includegraphics[width=0.4\textwidth]{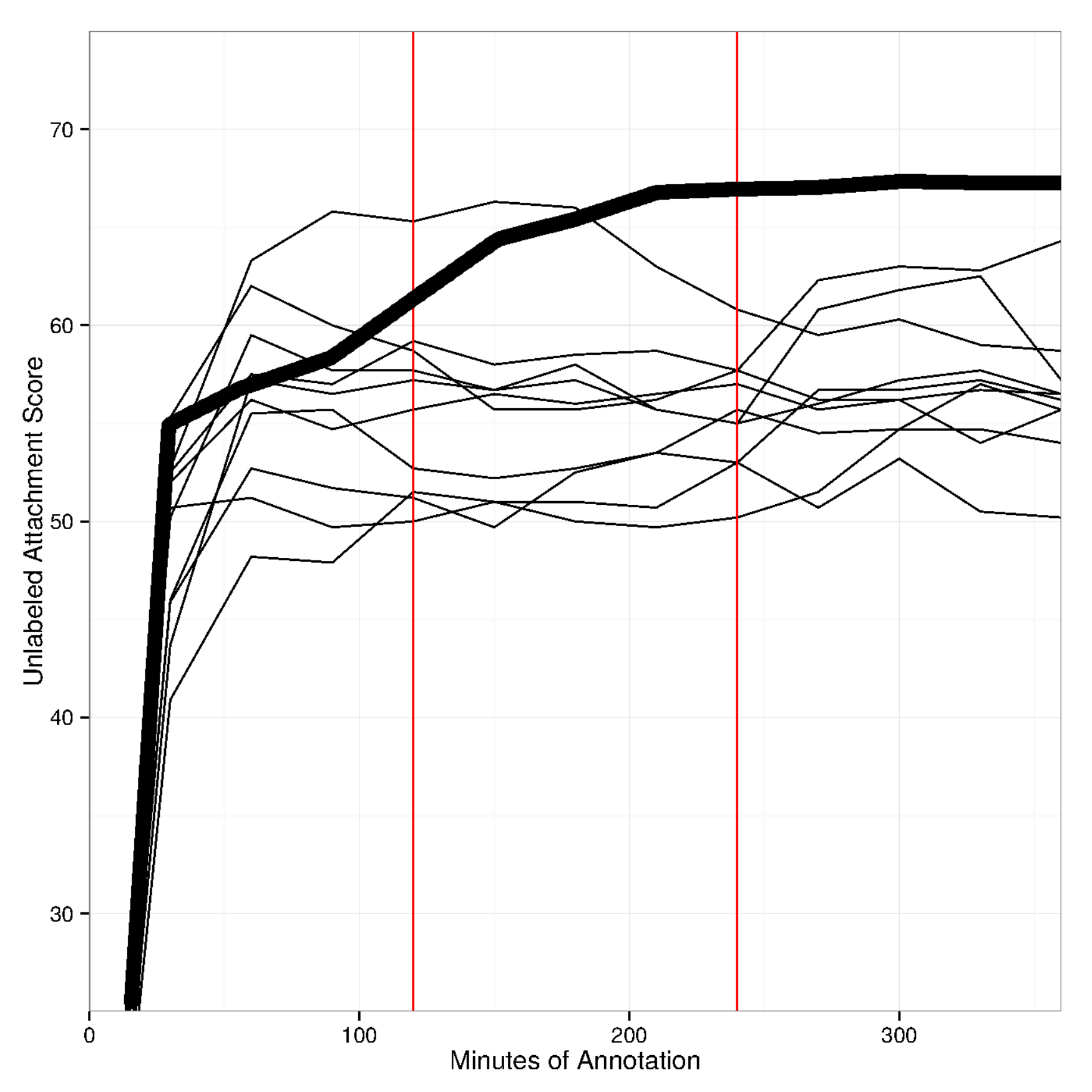}
    \caption{Learning curves for individual annotators (thin lines) and conglomerated training sets (thick line) over annotation time. Vertical lines indicate annotation session breaks. }
    \label{fig:learning-curves}
\end{figure}

The best method of all uses both partial annotations and language universals: ConvexMST-UG+GFL improves on ConvexMST-GFL for both languages and POS conditions. The impact of the combination is greater for English, which has less GFL annotation. Overall, these results show that this combination is robust to varying amounts of partial annotations: the UG constraints are strong on their own and provide a strong basis without annotations, they contribute when there are not many annotations available, and eventually become less essential (but remain unharmful) as more are provided.

It is important to recall that the GFL annotations have no specific conformity to the gold standards of either \textit{original} corpus. Our goal was to understand the overall behavior of different methods given the same free-wheeling, diverse annotations; it is likely that higher numbers would have been achieved had we guided annotators to use corpus conventions, or used full annotations provided by our annotators as the evaluation set. The former defeats the spirit of our exercise, and we did not have sufficient budget for the latter.

For Spanish, we also considered the performance of individual annotators alongside the full training set. The learning curves for individual annotators are shown in Figure \ref{fig:learning-curves}. There is substantial variation in the curves for the individual annotators; however, the curve based on the union of all annotations at each time step is smooth and is better than any individual past the three hour mark. One way to consider this is in terms of building an accurate parser quickly with multiple, diverse annotators, where wall clock time matters. Another way is to consider robustness with respect to possibly bad annotators. The next obvious steps would be to use active learning and to detect disagreement in annotators to either drop some or intervene to improve their quality. (Again, keep in mind that we are considering a ``cold start'' to this process, so there can be no gold standard for checking annotator quality.)

\paragraph{Comparison to Full Annotation}

To this point, all performance comparisons have been between different parse feature sets; we have demonstrated that the GFL features are complimentary to the UG features, and that when standing alone the GFL features are stronger than the UG features. The question of whether it might be more effective to simply have annotators produce full annotations is not addressed by these comparisons. To answer this question, we had our most experienced annotator fully annotate the same section that the other annotators did partially.
Producing these full annotations required roughly 13 hours of time from the single expert annotator. In comparison, the other annotators were able to partially annotate the same section in roughly two hours each -- a total of 24 hours. However, the theoretical wall clock time of the group of annotators could be as low as two hours if the sessions were run in parallel. These different training sets were once again used to train ConvexMST models that were evaluated on a held out test set. Table \ref{tab:full-vs-partial} contains the results of this experiment, demonstrating that the group of inexperienced annotators producing partial annotations was able to achieve similar performance levels to the single annotator producing full annotations.
It should be noted that this comparison does not weight the results using the extrinsic costs associated with the production of the training data. In a real-world environment, the expert annotator would likely be more expensive than the inexperienced annotators, and possibly all of them combined (especially in a crowd-sourcing scenario). This makes the performance per unit cost for partial annotators even higher than Table \ref{tab:full-vs-partial} indicates. See Section \ref{ssec:cost} for discussion and modeling of these extrinsic cost effects.

\begin{table}
  \begin{tabular}{l|c|c}
    Feature Set & Partial Annotations & Full Annotations \\ \hline
    UG          & 56.9                & 58.8             \\
    GFL         & 61.2                & 62.8             \\
    GFL+UG      & \textbf{63.2}                & \textbf{66.6}             \\
  \end{tabular}
  \caption {Comparison between full and partial annotations, 10 or fewer words, using predicted POS tags.}
  \label{tab:full-vs-partial}
\end{table}

\subsection{Longer Sentences}

We also evaluated ConvexMST with longer sentences: those with 20 words or less. For this, the right-branching baseline is 25.8\%. When using all the annotations on the common set for all annotators, the scores for ConvexMST with UG, GFL, and GFL+UG are 47.6\%, 54.4\%, and 55.3\%, respectively. The values are worse than for shorter sentences, as expected, but the pattern observed in Table \ref{tab:parsing-results} still holds: GFL annotations best UG alone, and their combination is the best of all.

\subsection{Discussion \& Error Analysis}

\paragraph{POS-Tagging Impact}

\begin{figure*}[!ht]
  \centering
  \begin{subfigure}[b]{0.48\textwidth}
    \includegraphics[width=\textwidth]{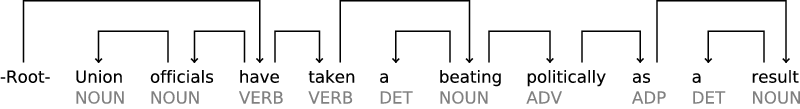}
    \caption{Gold Tags}
    \label{fig:gold_parse}
  \end{subfigure}
  \quad
  \begin{subfigure}[b]{0.48\textwidth}
    \includegraphics[width=\textwidth]{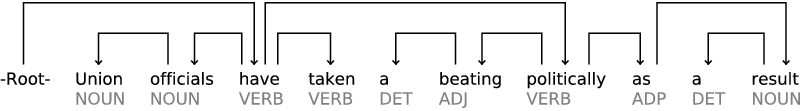}
    \caption{Predicted Tags}
    \label{fig:pred_parse}
  \end{subfigure}
  \caption{Differences in parsing results due to minimal POS tagging errors.}
  \label{fig:pos-parsing-error}
\end{figure*}

We thought it important to consider imperfect POS-taggings because this entire framework is based off of the assumption that the user is working from essentially no pre-existing resources. Assuming the availability of gold-standard POS tags is antithetical to this idea, and is one way in which direct supervision can show up in otherwise unsupervised (or indirectly supervised) systems. 

Many tagger errors are not likely to cause major problems during parsing; for instance mislabeling pronouns as nouns, or adverbs as adjectives, is unlikely to lead to major structural issues. However, more unlikely errors can cause more dramatic effects, as shown in Figure~\ref{fig:pos-parsing-error}. Here, the phrase `beating politically' (gold tags `\textsc{noun adv}') is mis-tagged as `\textsc{adj verb}', leading to the attachment of `politically' to the root word and the reorganization of a substantial chunk of the sentence.

\paragraph{Weighting Constraint Violations}

For feature sets with both GFL and UG-based constraints, a weighting factor can bias the parser towards being more likely to respect either GFL or UG constraints. We experimented with this, and found that for the datasets we considered, the best results were obtained when we weighted violations of GFL constraints as worse than violations of UG constraints. This result is not entirely unexpected given the relative performances of the constraints on their own, but it provides more evidence that direct supervision even in small amounts can beat indirect supervision.

\section{Conclusion}

We have shown that human-sourced partial annotations can be exploited to learn effective dependency parsers in short period of time. The ConvexMST method we adapt from Grave and Elhadad easily combines constraints from both language universals and partial annotations, providing greater robustness from starting annotation until one runs out of budget or time. We demonstrate this with actual annotations produced for English and Spanish, using annotators with a range of experience.

Overall, we present a case for working in realistic settings by paying close attention to the various sources of annotation and tracking the real costs associated with that supervision. We believe that over-reliance on creeping supervision of this type may lead to an inaccurate picture of the cross-lingual and low-resource applicability of various models, and are encouraged by recent work on character-based models by Gillick et al. \shortcite{bytenlp} and Ballesteros et al \shortcite{ballesteros:2015}, among others. Their work shows viable models can be produced without relying on having annotations a priori, but rather learning representations on the fly that need not conform to any one set of standards.

\section*{Acknowledgments}
Supported by the U.S. Army Research Office under grant number W911NF-10-1-0533. Any opinions, findings, and conclusions or recommendations expressed in this material are those of the author(s) and do not necessarily reflect the view of the U.S. Army Research Office.

\nocite{*}

\bibliographystyle{emnlp2016}
\bibliography{partial-convex}

\end{document}